\documentclass[10pt,twocolumn,letterpaper]{article}

\usepackage[pagenumbers]{cvpr} 

\usepackage{graphicx}
\usepackage{amsmath}
\usepackage{amssymb}
\usepackage{booktabs}
\usepackage{xcolor}
\usepackage{multirow}
\usepackage[noend]{algpseudocode}
\usepackage{algorithmicx,algorithm}
\usepackage{array}
\usepackage{bbm}

\usepackage[pagebackref,breaklinks,colorlinks]{hyperref}

\usepackage[capitalize]{cleveref}
\crefname{section}{Sec.}{Secs.}
\Crefname{section}{Section}{Sections}
\Crefname{table}{Table}{Tables}
\crefname{table}{Tab.}{Tabs.}

\begin{document}

\title{Self-Supervised Consistent Quantization for Fully Unsupervised Image Retrieval}

\author{Guile Wu$^{*}$\qquad
Chao Zhang$^{1}$\qquad
Stephan Liwicki$^{1}$\qquad
\\
$^{1}$Toshiba Europe Limited, Cambridge, UK\\
{\tt\small guile.wu@outlook.com, \{czhang, stephan.liwicki\}@crl.toshiba.co.uk}
}
\maketitle

\newcommand\blfootnote[1]{%
  \begingroup
  \renewcommand\thefootnote{}\footnote{#1}%
  \addtocounter{footnote}{-1}%
  \endgroup
}
\blfootnote{* Work done when G. Wu was an intern at Toshiba Europe Limited.}

\begin{abstract}
Unsupervised image retrieval aims to learn an efficient retrieval system without expensive data annotations, but most existing methods rely heavily on handcrafted feature descriptors or pre-trained feature extractors.
To minimize human supervision, recent advance proposes deep fully unsupervised image retrieval aiming at training a deep model from scratch to jointly optimize visual features and quantization codes.
However, existing approach mainly focuses on instance contrastive learning without considering underlying semantic structure information, resulting in sub-optimal performance.
In this work, we propose a novel self-supervised consistent quantization approach to deep fully unsupervised image retrieval, which consists of part consistent quantization and global consistent quantization.
In \emph{part consistent quantization}, we devise part neighbor semantic consistency learning with codeword diversity regularization.
This allows to discover underlying neighbor structure information of sub-quantized representations as self-supervision.
In \emph{global consistent quantization}, we employ contrastive learning for both embedding and quantized representations and fuses these representations for consistent contrastive regularization between instances.
This can make up for the loss of useful representation information during quantization and regularize consistency between instances.
With a unified learning objective of part and global consistent quantization, our approach exploits richer self-supervision cues to facilitate model learning.
Extensive experiments on three benchmark datasets show the superiority of our approach over the state-of-the-art methods.
\end{abstract}

\section{Introduction}
Image retrieval is a fundamental task in computer vision, aiming to find images that are visually similar to a given query image from a large database.
To reduce computational cost and improve storage efficiency, approximate nearest neighbor search~\cite{indyk1998approximate} has been widely used, where hashing~\cite{charikar2002similarity,shen2020auto,yuan2020central,yang2019distillhash} and product quantization~\cite{jegou2010product,jang2021self,klein2019end,wang2022contrastive} are two most representative directions.
Hashing maps real-value embeddings to binary codes for efficient retrieval, while product quantization divides real-value data space into disjoint partitions to quantize embeddings for efficient retrieval.
In the past decade, the incredible success of deep learning in computer vision has brought a great breakthrough to deep supervised hashing~\cite{cao2017hashnet,yuan2020central,li2017deep} and deep supervised quantization~\cite{yu2020product,yue2016deep,klein2019end} based image retrieval.
Although deep supervised image retrieval methods have shown outstanding performance, the reliance of expensive label annotations of training data hinders their applications in label-limited scenarios.

On the other hand, unsupervised image retrieval is capable of learning an efficient retrieval system without using labeled training data.
Traditional unsupervised image retrieval methods~\cite{heo2012spherical,liu2014discrete,babenko2014additive} utilize handcrafted descriptors to extract embeddings of input images and adopt unsupervised hashing or quantization approaches to efficient retrieval.
Recent deep unsupervised image retrieval methods~\cite{li2021deep,yang2019distillhash,wang2022contrastive} resort to ImageNet~\cite{russakovsky2015imagenet} pre-trained deep neural networks for feature extraction and incorporate deep hashing or deep quantization techniques into a deep model for optimization.
However, these methods either rely on human supervision for devising effective handcrafted feature descriptors or supervised pre-trained ImageNet backbone networks.
To minimize human supervision, deep fully unsupervised image retrieval is recently proposed in~\cite{jang2021self},
which aims to train a deep model from scratch to jointly optimize visual features and codes for efficient image retrieval.
Despite the self-supervised product quantization approach introduced in~\cite{jang2021self} has shown promising performance, it only focuses on instance cross quantized contrastive learning without considering underlying semantic structure information, resulting in sub-optimal performance.

\begin{figure*}[t]
\centering
   \includegraphics[width=0.95\linewidth]{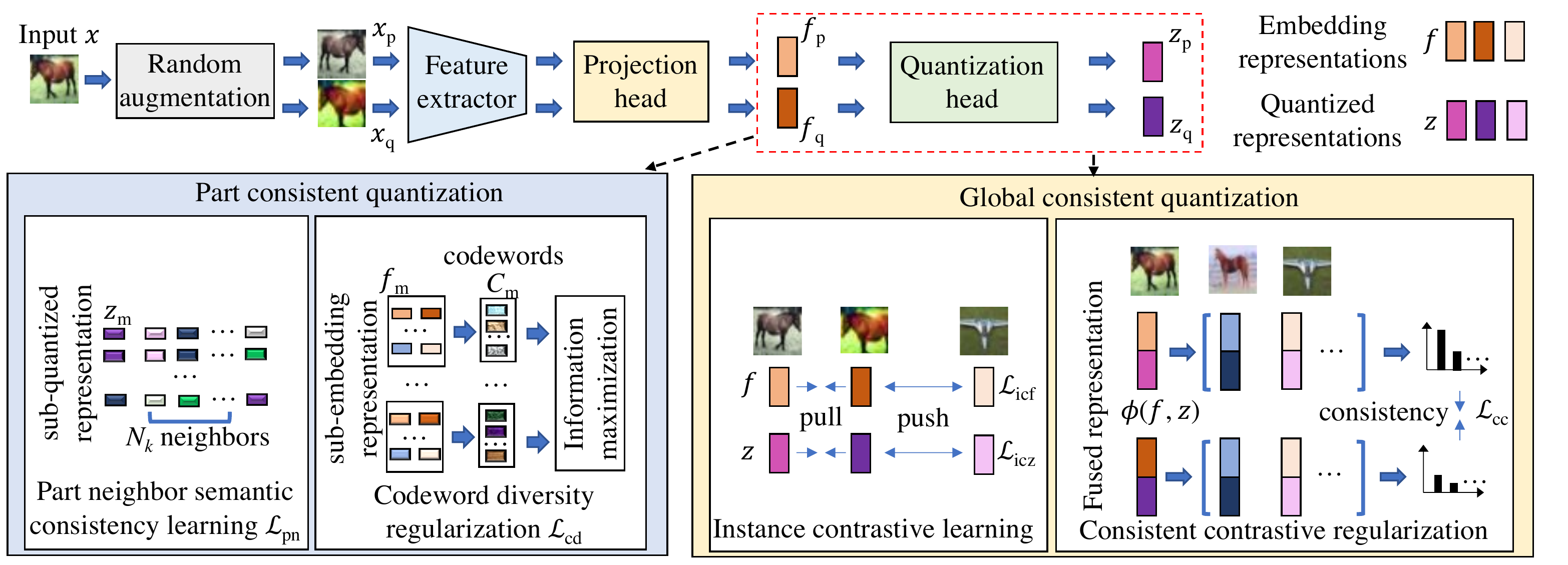}
   \vspace{-0.3cm}
   \caption{
   An overview of the proposed Self-Supervised Consistent Quantization (SSCQ) approach to deep fully unsupervised image retrieval.
   Part consistent quantization discovers part neighbor affinity as self-supervision, while global consistent quantization learns instance affinity as self-supervision, which together are formulated into a unified learning objective for model optimization.
   }
   \label{fig:overview}
\end{figure*}

In this work, we introduce a novel Self-Supervised Consistent Quantization (SSCQ) approach to deep fully unsupervised image retrieval.
We propose to explore richer part and global self-supervision for learning underlying semantic structure information to facilitate model learning.
An overview of the proposed approach is depicted in Fig.~\ref{fig:overview}.
It is known that neighbors in embedding space usually share useful semantic information~\cite{wu2020tracklet,huang2019unsupervised}, since the quantization process in product quantization is akin to clutering, the sub-quantized representations should also share underlying neighbor semantic structure information.
In light of this, with a contrastive quantization baseline~\cite{jang2021self}, we devise part neighbor semantic consistency learning with codeword diversity regularization~\cite{krause2010discriminative,klein2019end} to encourage semantic consistency of neighbors of sub-quantized representations.
This resolves the limitation of the existing contrastive quantization methods~\cite{jang2021self,wang2022contrastive} that fail to learn underlying semantic structure information.
We term this process \emph{part consistent quantization}.
Meanwhile, since it is inevitable that the reconstruction in the quantization process will lose useful information of embedding representations, we apply instance contrastive learning for both embedding and quantized representations to simultaneously optimize visual features and codes so as to make up for the loss.
Besides, we fuse the embedding and quantized representations and apply consistent contrastive regularization to regularize consistency between instances.
We term this process \emph{global consistent quantization}.
Both global and part consistent quantization are formulated into a unified learning objective to explore richer self-supervision for deep fully unsupervised image retrieval.
To evaluate the effectiveness of the proposed approach, we conduct extensive experiments on three benchmark datasets, namely CIFAR-10~\cite{krizhevsky2009learning}, NUS-WIDE~\cite{chua2009nus} and FLICKR25K~\cite{huiskes2008mir}.
Experimental results show the superiority of our approach over the state-of-the-art methods.

Our \textbf{contributions} are:
We propose a novel self-supervised consistent quantization approach consists of global consistent quantization and part consistent quantization to deep fully unsupervised image retrieval.
In part consistent quantization, we use part neighbor semantic consistency learning with codeword diversity regularization, which helps to learn underlying neighbor semantic structure information of sub-quantized representations.
In global consistent quantization, we employ contrastive learning for both embedding and quantized representations and fuse these representations for consistent contrastive regularization, which helps to make up for the loss of useful representation information during quantization and regularize consistency between instances.

\section{Related Work}
\paragraph{Unsupervised Image Retrieval.}
Most traditional image retrieval methods are originally designed for unsupervised learning scenarios where no labeled training data are used for model learning.
The common practice usually consists of two disjoint steps.
The first step is to use handcrafted descriptors, such as GIST~\cite{oliva2001modeling} and SIFT~\cite{lowe2004distinctive}, to extract features of input images, while the second step is to employ binary hashing~\cite{charikar2002similarity,weiss2008spectral,salakhutdinov2009semantic} or product quantization~\cite{jegou2010product,ge2013optimized,babenko2014additive} to transform embedding space into Hamming space or a Cartesian product of subspaces for efficient image retrieval. 
In the past decade, deep learning based methods have dominated the field of image retrieval.
Supervised deep hashing~\cite{yuan2020central,li2017deep} or deep product quantization~\cite{yu2020product,klein2019end} based image retrieval methods have shown notably better performance than the traditional counterparts.
However, these supervised methods are limited by the availability of labeled data for model training.
To resolve this problem, researchers have resorted to deep unsupervised learning for image retrieval.
One of the most popular direction is deep unsupervised hashing~\cite{yang2019distillhash,li2021deep}.
An ImageNet~\cite{russakovsky2015imagenet} pre-trained deep neural network is usually used as the feature encoder and then hashing layers are inserted into the model to learn discriminative binary codes using unlabeled data.
Despite outstanding performance has been achieved, deep unsupervised hashing relies heavily on the pre-trained feature encoder and can only learn restricted binary codes, limiting its ability for distinguishing visually similar but semantic dissimilar data.
Since product quantization is capable of learning continuous representations for efficient image retrieval, deep unsupervised product quantization~\cite{wang2022contrastive,jang2021self} is recently introduced for unsupervised image retrieval.
In~\cite{wang2022contrastive}, soft quantization~\cite{yu2020product} is combined with contrastive learning~\cite{chen2020simple} and code memory to learn a retrieval system in an unsupervised manner.
However, \cite{wang2022contrastive} still requires using a pre-trained model as the feature extractor and most layers are not optimized during model training.
To minimize human supervision, deep fully unsupervised image retrieval is introduced in~\cite{jang2021self}, which aims to train a deep model from scratch for efficient image retrieval.
In~\cite{jang2021self}, a cross-quantized contrastive learning framework is proposed to jointly optimize visual features and codes for unsupervised image retrieval.

Our work belongs to deep unsupervised product quantization and focuses on deep fully unsupervised image retrieval without data label annotation nor supervised pre-trained backbone models.
We propose a novel self-supervised consistent quantization approach to discover richer self-supervision to facilitate model optimization.
We devise part consistent quantization to discover underlying neighbor semantic structure information and global consistent quantization to learn affinity between instances.
This differs from~\cite{jang2021self} that only uses contrastive learning to optimize cross-quantized representations or \cite{wang2022contrastive} that requires a pre-trained model as the feature encoder as well as an additional code memory.

\paragraph{Self-Supervised Representation Learning.}
In recent years, self-supervised/unsupervised representation learning has made great progress.
The common practice is devising different pretext tasks, such as predicting image rotation~\cite{komodakis2018unsupervised} and solving jigsaw puzzles~\cite{noroozi2016unsupervised}, to generate self-supervision information to facilitate unsupervised representation learning.
Recently, contrastive learning~\cite{chen2020simple,he2020momentum,wu2018unsupervised,wei2021co2} has become one of the most popular and powerful paradigms for unsupervised representation learning.
It usually applies strong random augmentation to each input image to generate positive counterparts and employs a contrastive loss to pull the positives closer and push the negatives apart, where different instances are considered as negatives.
In deep fully unsupervised image retrieval, \cite{jang2021self} introduces a contrastive quantization framework which shows promising performance compared with conventional unsupervised image retrieval methods even without supervised pre-training.
Our approach introduces self-supervised consistent quantization to discover underlying neighbor semantic structure information from sub-quantized representations and to learn affinity between instance so as to facilitate model learning in deep fully unsupervised image retrieval.

\section{Methodology}
\paragraph{Problem Statement.}
In this work, we study deep fully unsupervised image retrieval~\cite{jang2021self}, where no labeled training data nor pre-trained models are available.
Given an unlabeled training set $\mathcal{X}{=}\{x_i\}_{i=1}^{N}$ with $N$ unlabeled samples, our task is to learn a model to encode $x_i$ into a $L$-bit code $B_i{=}\{b_i^j\}_{j=1}^L$, where $b_i^j{\in}\{0, 1\}$, for efficient image retrieval.
During inference, the similarity between query and database samples are measured based the learned representations and codes so as to realize efficient retrieval.

\subsection{Approach Overview}
An overview of the proposed Self-Supervised Consistent Quantization (SSCQ) approach is depicted in Fig.~\ref{fig:overview}.
In each training mini-batch $\{x_i\}_{i=1}^{N_b}$, we apply strong random data augmentation~\cite{chen2020simple} on each input sample to generate two augmented views, so we have ${2N_b}$ augmented samples in each mini-batch.
Then, we extract a $D$-dimensional embedding representations $f_i$ of an augmented input $x_i$ and further quantize $f_i$ into a $D$-dimensional quantized representation $z_i$.
To construct the model training objective, we employ part and global consistent quantization.
In part consistent quantization, we compute a part neighbor semantic consistency learning loss $\mathcal{L}_{pn}$ for each sub-quantized representation and use entropy maximization $\mathcal{L}_{cd}$ to regularize the diversity of codewords.
Meanwhile, in global consistent quantization, we compute instance contrastive learning loss $\mathcal{L}_{icf}$ for embedding representations $\{f_i\}_{i=1}^{2N_b}$ and $\mathcal{L}_{icz}$ for quantized representations $\{z_i\}_{i=1}^{2N_b}$ respectively, and compute consistent contrastive regularization $\mathcal{L}_{cc}$ with fused representations $\{\phi(f_i, z_i)\}_{i=1}^{2N_b}$.
Therefore, the unified learning objective is formulated as:
\begin{equation}
\label{eq:learning_objective}
\mathcal{L} = \mathcal{L}_{icz} + {\lambda}_{pn}\mathcal{L}_{pn} + {\lambda}_{cd}\mathcal{L}_{cd} + \mathcal{L}_{icf} + {\lambda}_{cc}\mathcal{L}_{cc},
\end{equation}
where ${\lambda}_{pn}$, ${\lambda}_{cd}$ and ${\lambda}_{cc}$ are weighting parameters.
In the following section, we drop the subscript $i$ (e.g., let $x$ denote an input $x_i$) if the context is clear.

\subsection{Self-Supervised Consistent Quantization}
\paragraph{A Baseline with Contrastive Quantization.}
We employ a contrastive quantization baseline model~\cite{jang2021self} for deep fully unsupervised image retrieval.
As shown in Fig.~\ref{fig:overview}, with each augmented input sample $x$, a feature extractor is used to extract feature of $x$ and a projection head is employed to map the learned feature into a $D$-dimensional embedding representation $f$.
The feature extractor is constructed with a deep convolutional neural network, while the projection head is a Multi-Layer Perceptron (MLP) where the first fully connected (FC) layer maps the feature to $512$-dimension following a ReLU layer and the second FC layer outputs a $D$-dimensional $f$.
Then, a quantization head is used to reconstruct $f$ into a $D$-dimensional  quantized representation $z$.
Suppose there are $M$ codebooks $\{C_m\}_{m=1}^{M}$ in the quantization head and each codebook is composed of $K$ codewords $C_m{=}\{c_{m,k}\}_{k=1}^K$, where $c_{m,k}{\in}\mathbb{R}^{M/D}$.
Following product quantization~\cite{jegou2010product,yu2020product}, $f$ is divided into $M$ disjoint sub-embedding representation $f_{m}{\in}\mathbb{R}^{M/D}$ and the codewords in the $m$-th codebook attempt to reconstruct the sub-embedding representation $f_{i,m}$.
Therefore, the embedding space is divided into a Cartesian product of $M$ subspaces $\{C_1{\times}C_2{\times}...{\times}C_M\}$, and codewords in the $m$-th codebook are considered as distinct clustered centroids of the $m$-th sub-embedding representations of all samples.
This allows to cluster visually similar sub-embedding representations into the same codeword for efficient similarity measurement.
To train the feature extractor, the projection head and the quantization head in an end-to-end manner, soft quantization~\cite{yu2020product} is employed for training, so the sub-quantized representation $z_{m}$ is defined as:
\begin{equation}
\label{eq:soft_quantization}
z_{m} = \sum_{k=1}^{K}\frac{exp(d(f_{m}, c_{m,k})/\tau_{sq})}{{\sum_{j=1}^K}exp(d(f_{m}, c_{m, j})/\tau_{sq})}c_{m,k},
\end{equation}
where $d(f_{m}, c_{m,k}) = -\|f_{m}-c_{m,k}\|^2_2$ is the squared Euclidean distance, and $\tau_{sq}$ is a temperature parameter, $z$ is the concatenation of $z_{m}$.

To optimize the model, an instance contrastive learning loss $L_{icz}$~\cite{chen2020simple} is computed to pull $z$ closer to its positive (the reconstructed representation of augmented counterparts from the same input) and push its negatives apart (the reconstructed representations of other augmented samples of different inputs), as:
\begin{equation}
\label{eq:instance_contrastive_z}
\mathcal{L}_{icz}=-{log}\frac{exp(s(z_{}, z_{}^+)/\tau_{ic})}{\sum_{j=1}^{2N_b}\mathbbm{1}_{[z_j{\neq}z]}exp(s(z_{}, z_{j})/\tau_{ic})},
\end{equation}
where $z_{}$ and $z_{}^+$ are positive pairs of the same input, $s(z_{}, z_{}^+) = z_{}^Tz_{}^+/(\|z_{}\|\|z_{}^+\|)$ is cosine similarity between $z_{}$ and $z_{}^+$, $\mathbbm{1}_{[z_j{\neq}z]}$ denotes an indicator function, and $\tau_{ic}$ is a temperature parameter.

With the baseline model, \cite{jang2021self} reformulates a cross quantized contrastive learning loss to learn the correlation between the embedding representation $f$ and the quantized representation $z$.
However, it only considers instance contrastive learning and largely ignores the underlying semantic structure information.
To resolve this problem, we propose to use part and global consistent quantization to explore richer self-supervision cues for model optimization.

\paragraph{Part Consistent Quantization.}
The quantization process in the quantization head is akin to clustering by learning distinct codewords as cluster centroids but with learnable parameters for end-to-end model training.
Thus, each sub-quantized representation $z_{m}$ inherently encodes part neighbor semantic structure information, which can be exploited as auxiliary self-supervision cues to facilitate model learning.
In light of this, we use a part neighbor semantic consistency learning loss $\mathcal{L}_{pn}$ to pull each sub-quantized representation $z_{m}$ closer to its similar sub-quantized representations and push $z_{m}$ away from dissimilar ones, as:
\begin{equation}
\label{eq:part_discriminative}
\mathcal{L}_{pn}=-\frac{1}{M}\sum_{m=1}^{M}{log}\frac{\sum_{n=1}^{N_k}{exp(s(z_{m}, z_{m,n}^{-})/\tau_{pn})}}{\sum_{j=1}^{2Nb-2}{exp(s(z_{m}, z_{m,j}^{-})/\tau_{pn})}},
\end{equation}
where $\tau_{pn}$ is a temperature parameter, and $\{z_{m,n}^{-}\}_{n=1}^{N_k}$ are the top $N_k$ part neighbors of $z_{m}$ which are obtained by computing the similarity between $z_{m}$ and its negative sub-quantized representations $z_{m}^-$.
Here, $\mathcal{L}_{pn}$ differs from existing unsupervised neighbor discovery methods~\cite{huang2019unsupervised,yang2019patch,wu2020tracklet} in that $\mathcal{L}_{pn}$ mines neighbor affinity between sub-quantized representations $z_{m}$ and its negatives $z_{m,n}^{-}$ to facilitate quantization, instead of progressively exploring anchored neighbors~\cite{huang2019unsupervised} or maintaining a patch/tracklet memory bank~\cite{yang2019patch,wu2020tracklet}.

Besides, to encourage diverse codeword distribution, we compute the similarity between sub-embedding representations and codewords in each codebook and encourage the mean probability distribution to be diverse, as:
\begin{equation}
\label{eq:codeword_diversity}
\mathcal{L}_{cd} = \frac{1}{M}\sum_{m=1}^{M}\sum_{k=1}^{K}\hat{p}_{m,k}{\cdot}{log}(\hat{p}_{m,k}),
\end{equation}
where $\hat{p}_{m,k}{=}\frac{1}{2N_b}\sum_{i}^{2N_b}\frac{exp(s(f_{i,m}, c_{m,k}))}{\sum_{t=1}^{K}exp(s(f_{i,m}, c_{m,t}))}$ is the mean output probability over all samples in a mini-batch.
Note that similar codeword diversity regularization has been used in previous quantization method~\cite{klein2019end}, but here $\mathcal{L}_{cd}$ in our approach is an auxiliary term based on entropy maximization~\cite{krause2010discriminative,liang2020we} for unsupervised part consistent quantization and is not directly computed using soft quantization code.

\paragraph{Global Consistent Quantization.}
Previous contrastive quantization based methods~\cite{jang2021self,wang2022contrastive} mainly use contrastive learning for quantized or cross-quantized representations.
However, it is inevitable that quantized representations lose some useful embedding representation information during reconstruction in the quantization process.
This issue potentially hinders model optimization when the feature extractor is not pre-trained and not fixed in an end-to-end trainable framework.
To address this issue, as shown in Fig.~\ref{fig:overview}, in addition to $\mathcal{L}_{icz}$, we also compute an instance contrastive learning loss $\mathcal{L}_{icf}$ for the embedding representations $f$, as:
\begin{equation}
\label{eq:instance_contrastive_f}
\mathcal{L}_{icf}=-{log}\frac{exp(s(f_{}, f_{}^+)/\tau_{ic})}{\sum_{j=1}^{2N_b}\mathbbm{1}_{[f_j{\neq}f]}exp(s(f_{}, f_{j})/\tau_{ic})}.
\end{equation}
With Eqs.\eqref{eq:instance_contrastive_z} and \eqref{eq:instance_contrastive_f}, we can simultaneously optimize the reconstructed quantized representations and the original embedding representations to jointly learn visual features and codes.

Furthermore, recent advance in unsupervised representation learning~\cite{li2021prototypical,wei2021co2} indicates that exploring the affinity between positive and negative instances in contrastive learning is beneficial to model optimization for better generalization.
To further facilitate unsupervised quantization, we apply a consistent contrastive regularization~\cite{wei2021co2} $\mathcal{L}_{cc}$ to the global consistent quantization.
Directly applying $\mathcal{L}_{cc}$ on model outputted quantized representations may not fully explore global semantic information.
Thus, we first fuse $f$ and $z$ in order to learn the potential correlation between embedding and quantized representations, i.e., generate the fused representation $\Phi(f, z)$ where $\Phi(\cdot, \cdot)$ is the fusion operation (e.g., concatenation or sum fusion).
Then, we compute the similarity $Q(j)$ between $\Phi(f, z)$ and its negatives $\{\Phi(f_{}^-, z_{}^-)_j\}_{j=1}^{2N_b-2}$ and the similarity $P(j)$ between $\Phi(f^+, z^+)$ and the same negatives, as:
\begin{equation}
\begin{aligned}
\label{eq:similarity_negative}
Q(l) = \frac{exp(s(\Phi(f, z), \Phi(f_{}^-, z_{}^-)_l)/\tau_{cc})}{\sum_{j=1}^{2N_b-2}{exp(s(\Phi(f, z), \Phi(f_{}^-, z_{}^-)_j)/\tau_{cc})}},\\
P(l) = \frac{exp(s(\Phi(f^+, z^+), \Phi(f_{}^-, z_{}^-)_l)/\tau_{cc})}{\sum_{j=1}^{2N_b-2}{exp(s(\Phi(f^+, z^+), \Phi(f_{}^-, z_{}^-)_j)/\tau_{cc})}},
\end{aligned}
\end{equation}
where $\tau_{cc}$ is a temperature parameter.
Since unlabeled samples in a training set can usually be categorized into multiple semantical categories where each category contains multiple samples, the similarity between an instance and its negatives potentially encodes the neighbor semantic structure information.
By regularizing the consistency of similarity probability distribution between an instance and its positive, we can encourage the model to output consistent predictions between different augmented instances from the same input.
Thus, using the symmetric Kullback-Leibler Divergence $D_{KL}$, $\mathcal{L}_{cc}$ is defined as:
\begin{equation}
\label{eq:consistent_contrastive}
\mathcal{L}_{cc}=\frac{1}{2}(D_{KL}(P\|Q)+D_{KL}(Q\|P)).
\end{equation}
Note that the fused representation $\Phi(f, z)$ is built on the instance contrastive learning of both $f$ and $z$, so if only $z$ is optimized with contrastive learning, then $\Phi(f, z)=z$.
Alternatively, we can compute cross consistent contrastive regularization in Eqs.\eqref{eq:similarity_negative} and \eqref{eq:consistent_contrastive} for embedding and quantized representations to encourage consistent predictions between them.

\paragraph{Summary.}
We summarize the training process of the proposed Self-Supervised Consistent Quantization in Algorithm~\ref{alg:sscq}.
In training, a unified learning objective (Eq.\eqref{eq:learning_objective}) based on part and global consistent quantization is formulated for model learning.
In inference, following the previous work~\cite{jang2021self,wang2022contrastive}, we use hard quantization to generate the $(M{\cdot}{log_2}K)$-bits code for each sample in the database by finding the most similar codeword $\{c_{m,k}\}_{k=1}^K$ in each codebook $\{C_m\}_{m=1}^{M}$ for each sub-embedding representation.
Then, we use asymmetric distance~\cite{jegou2010product} to measure the distance between each query sample and database samples.
Specifically, given a query image, we extract its embedding representation and divide it into $M$ sub-embedding representations.
Next, we compute the Euclidean distance between each sub-embedding representation and all codewords in all codebooks to set up a query-specific look-up table.
Finally, we can approximately calculate the distance between the query sample and each database sample by using the code to get the sub-vector distance from the query-specific look-up table and summing up.

\begin{algorithm}[t]
\caption{{Self-Supervised Consistent Quantization.}}
{\bf Input:} A baseline model, unlabeled training data $\mathcal{X}$
\begin{algorithmic}[1]
\For{sampled mini-batch $\{x_i\}_{i=1}^{N_b}$}
   \State Generate two augmented samples for each $x$
   \State Extract embedding representation $f$ of all samples
   \State Extract quantized representation $z$ of all samples
   \State Compute $\mathcal{L}_{icz}$ for $z$ (Eq.\eqref{eq:instance_contrastive_z})
   \State /* Part consistent quantization */
   \State Compute $\mathcal{L}_{pn}$ for $z_m$ (Eq.\eqref{eq:part_discriminative})
   \State Compute $\mathcal{L}_{cd}$ for $f_m$ and $c_{m,k}$ (Eq.\eqref{eq:codeword_diversity})
   \State /* Global consistent quantization */
   \State Compute $\mathcal{L}_{icf}$ for $f$ (Eq.\eqref{eq:instance_contrastive_f})
   \State Compute $\mathcal{L}_{cc}$ for fused $\phi(f,z)$ (Eq.\eqref{eq:consistent_contrastive})
   \State /* Unified learning objective */
   \State Optimize the model with $\mathcal{L}$ (Eq.\eqref{eq:learning_objective})
\EndFor
\State {\bf end for}
\end{algorithmic}
{\bf Output:} A trained model for image retrieval.
\label{alg:sscq}
\end{algorithm}

\section{Experiments}
\subsection{Dataset and Evaluation Protocol}
\noindent\textbf{Datasets.}
To evaluate the proposed self-supervised consistent quantization approach for fully unsupervised image retrieval, we conduct extensive experiments on three datasets, namely CIFAR-10~\cite{krizhevsky2009learning}, NUS-WIDE~\cite{chua2009nus} and FLICKR25K~\cite{huiskes2008mir}.
\textbf{CIFAR-10} consists of 60,000 images of 10 class, where each class has 5,000 images for training and 1,000 images for testing.
We use 1,000 images per class as the query set, while the remaining images are used as the training set and the retrieval database.
\textbf{NUS-WIDE} is a multi-label large-scale dataset with around 270,000 images of 81 categories.
We select images of the 21 most frequent categories for evaluation, where 100 images per categories are selected to form 21,000 images as the query set while the remaining images form the training set and the retrieval database.
\textbf{FLICKR25K} is a relatively small dataset with 25,000 images of 24 categories.
We randomly select 2,000 images as the query set while the remaining images are used as the training set and the retrieval database.

\vspace{0.2cm}
\noindent\textbf{Evaluation Metrics.}
Following~\cite{jang2021self,li2021deep,wang2022contrastive,shen2020auto}, we mainly employ mean Average Precision (mAP, \%) as the evaluation metric.
We use mAP@1000 for CIFAR-10 and mAP@5000 for NUS-WIDE and FLICKR25K, and report image retrieval results with $\{16, 32, 64\}$ bits codes.
Besides, we also report Precision-Recall curves (PR) and Precision curves with top-1000 returned samples (P@1000) at 32 bits codes.
On the multi-label NUS-WIDE and FLICKR25K, if a query image and a database image share at least one label, then they are defined as the true match~\cite{li2021deep,jang2021self}.

\vspace{0.2cm}
\noindent\textbf{Implementation Details.}
We implement our approach with Python and PyTorch.
Following~\cite{jang2021self}, we use the modified ResNet-18~\cite{he2016deep,jang2021self} as the backbone (feature extractor) for CIFAR-10 where the first convolutional layer is modified with small kernel size and stride to adapt to the small $32{\times}32$ input image size, and use standard ResNet-50~\cite{he2016deep} as the backbone for NUS-WIDE and FLICKR25K.
We use strong random augmentation~\cite{chen2020simple}, including random cropping, horizontal flipping, color jitter, gray scaling and Gaussian blur, to generate augmented samples.
The number of codewords in each codebook is fixed to $K{=}2^4$, the dimension of each codeword is fixed to $D/M{=}16$ and the number of codebook is varying as $M{=}\{4, 8, 16\}$, so we can generate $\{M{\cdot}log_2{K}\}{=}\{16, 32, 64\}$ bits codes for image retrieval. 
We use Adam~\cite{kingma2014adam} as the optimizer with the initial learning rate of $5e{-}4$ for CIFAR-10 and $2e{-}4$ for NUS-WIDE and FLICKR25K, and set the weight decay of $1e{-}5$.
We warm up the learning rate with $10$ epochs and decay it with the cosine decay schedule~\cite{loshchilov2016sgdr} without restart.
On CIFAR-10, we set the batch size $N_B{=}256$ with the original input image size of $32{\times}32$, while on NUS-WIDE and FLICKR25K, $N_B{=}128$ with the input image size of $224{\times}224$.
In part consistent quantization, we set $\lambda_{pn}{=}0.1$, $\lambda_{cd}{=}0.2$, $N_k{=}20$, $\tau_{pn}{=}0.5$.
In global consistent quantization, we set $\tau_{sq}{=}0.2$ and $\tau_{ic}{=}0.5$ following~\cite{jang2021self}, and use $\lambda_{cc}{=}0.4$ and $\tau_{cc}{=}0.2$.
In fully unsupervised image retrieval, we train our model from scratch without using ImageNet~\cite{russakovsky2015imagenet} pre-trained weights.
Note that despite our approach is devised for deep fully unsupervised image retrieval, it is compatible with the deep pre-trained unsupervised setting, so we also report SSCQ-p that employs an ImageNet pre-trained VGG16 model as the backbone.

\newcommand{\tabincell}[2]{\begin{tabular}{@{}#1@{}}#2\end{tabular}}

\begin{table}[t]
   \begin{center}
   \resizebox{0.95\columnwidth}{!}{
      \begin{tabular}{c|c|ccc}
         \hline
         Type&Method&16 bits&32 bits&64 bits\\
         \hline
         \hline
         \multicolumn{1}{c|}{\multirow{5}{*}{\tabincell{c}{Shallow +\\pre-trained}}}
         &LSH~\cite{charikar2002similarity} &13.2 &15.8 &16.7 \\
         &SpectralH~\cite{weiss2008spectral} &27.2 &28.5 &30.0\\
         &PQ~\cite{jegou2010product} &23.7 &25.9 &27.2\\
         &ITQ~\cite{gong2012iterative} &30.5 &32.5 &34.9\\
         &OPQ~\cite{ge2013optimized} &29.7 &31.4 &32.3\\
         \hline
         \multicolumn{1}{c|}{\multirow{9}{*}{\tabincell{c}{Deep\\pre-trained\\unsupervised}}}
         &DeepBit~\cite{lin2016learning} &22.0 &24.9 &27.7\\
         &SAH~\cite{do2017simultaneous} &41.8 &45.6 &47.4\\
         &GreedyHash~\cite{su2018greedy} &44.8 &47.3 &50.1\\
         &SSDH~\cite{yang2018semantic} &36.2 &40.2 &44.0 \\
         &TBH~\cite{shen2020auto} &53.2 &57.3 &57.8 \\
         &CIBHash~\cite{qiu2021unsupervised} &59.4 &63.7 &65.2\\
         &Bi-half~\cite{li2021deep} &56.1 &57.6 &59.5\\
         &MeCoQ~\cite{wang2022contrastive}  &68.2 &69.7 &71.1\\
         &SSCQ-p (ours) &\textbf{76.1} &\textbf{76.8} &\textbf{78.1}\\
         \hline
         \multicolumn{1}{c|}{\multirow{5}{*}{\tabincell{c}{Deep fully\\unsupervised}}}
         &SGH~\cite{dai2017stochastic} &43.5 &43.7 &43.3\\
         &HashGAN~\cite{dizaji2018unsupervised} &44.7 &46.3 &48.1\\
         &BinGAN~\cite{zieba2018bingan} &47.6 &51.2 &52.0\\
         &SPQ~\cite{jang2021self} &76.8 &79.3 &81.2\\
         &SSCQ (ours) &\textbf{78.3} &\textbf{81.3} &\textbf{82.9}\\
         \hline
      \end{tabular}
   }
   \end{center}
\vspace{-0.3cm}
\caption{Comparison with the classic and state-of-the-art unsupervised methods on CIFAR-10 in terms of mAP (\%).
Some results are cited from~\cite{jang2021self,wang2022contrastive}.
}
\label{table:sota_cifar}
\end{table}

\begin{table}[t]
   \begin{center}
   \resizebox{0.95\columnwidth}{!}{
      \begin{tabular}{c|c|ccc}
         \hline
         Type&Method&16 bits&32 bits&64 bits\\
         \hline
         \hline
         \multicolumn{1}{c|}{\multirow{5}{*}{\tabincell{c}{Shallow +\\pre-trained}}}
         &LSH~\cite{charikar2002similarity} &38.5 &41.4 &43.9 \\
         &SpectralH$\dagger$~\cite{weiss2008spectral} &48.9 &53.0 &62.7 \\
         &PQ~\cite{jegou2010product} &65.4 &67.4 &68.6\\
         &ITQ$\dagger$~\cite{gong2012iterative} &68.0 &70.9 &72.8 \\
         &OPQ~\cite{ge2013optimized} &65.7 &68.4 &69.1\\
         \hline
         \multicolumn{1}{c|}{\multirow{7}{*}{\tabincell{c}{Deep\\pre-trained\\unsupervised}}}
         &DeepBit~\cite{lin2016learning} &39.2 &40.3 &42.9 \\
         &GreedyHash~\cite{su2018greedy} &63.3 &69.1 &73.1 \\
         &SSDH~\cite{yang2018semantic} &58.0 &59.3 &61.0 \\
         &CIBHash$\dagger$~\cite{qiu2021unsupervised} &79.5 &81.2 &81.7 \\
         &Bi-half~\cite{li2021deep} &76.9 &78.3 &79.9 \\
         &MeCoQ$\dagger$~\cite{wang2022contrastive}  &77.2 &81.5 &82.3 \\
         &SSCQ-p (ours) &\textbf{80.3} &\textbf{81.9} &\textbf{82.6} \\
         \hline
         \multicolumn{1}{c|}{\multirow{5}{*}{\tabincell{c}{Deep fully\\unsupervised}}}
         &SGH~\cite{dai2017stochastic} &59.3 &59.0 &60.7 \\
         &HashGAN~\cite{dizaji2018unsupervised} &68.4 &70.6 &71.7 \\
         &BinGAN~\cite{zieba2018bingan} &65.4 &70.9 &71.3 \\
         &SPQ$\dagger$~\cite{jang2021self} &75.7 &79.4 &80.2 \\
         &SSCQ (ours) &\textbf{78.7} &\textbf{79.9} &\textbf{80.8} \\
         \hline
      \end{tabular}
   }
   \end{center}
\vspace{-0.3cm}
\caption{Comparison with the classic and state-of-the-art unsupervised methods on NUS-WIDE in terms of mAP (\%).
Some results are cited from~\cite{wang2022contrastive}. $\dagger$ Reproduced results.
}
\label{table:sota_nus}
\end{table}

\begin{table}[t]
   \begin{center}
   \resizebox{0.95\columnwidth}{!}{
      \begin{tabular}{c|c|ccc}
         \hline
         Type&Method&16 bits&32 bits&64 bits\\
         \hline
         \hline
         \multicolumn{1}{c|}{\multirow{3}{*}{\tabincell{c}{Shallow +\\pre-trained}}}
         &LSH~\cite{charikar2002similarity} &58.8 &60.4 &64.2 \\
         &SpectralH~\cite{weiss2008spectral} &59.2 &60.6 &63.2 \\
         &ITQ~\cite{gong2012iterative} &68.4 &69.5 &70.3 \\
         \hline
         \multicolumn{1}{c|}{\multirow{5}{*}{\tabincell{c}{Deep\\pre-trained\\unsupervised}}}
         &GreedyHash~\cite{su2018greedy} &70.5 &72.3 &75.1 \\
         &SSDH~\cite{yang2018semantic} &78.7 &79.4 &79.5 \\
         &CIBHash~\cite{qiu2021unsupervised} &77.0 &78.5 &79.8 \\
         &Bi-half~\cite{li2021deep} &81.1 &82.4 &\textbf{82.9} \\
         &MeCoQ~\cite{wang2022contrastive}  &80.4 &81.7 &81.7 \\
         &SSCQ-p (ours) &\textbf{81.9} &\textbf{82.6} &82.8 \\
         \hline
         \multicolumn{1}{c|}{\multirow{2}{*}{\tabincell{c}{Deep fully\\unsupervised}}}
         &SPQ~\cite{jang2021self} &71.8 &74.0 &74.5 \\
         &SSCQ (ours) &\textbf{73.8} &\textbf{75.9} &\textbf{76.7}\\
         \hline
      \end{tabular}
   }
   \end{center}
\vspace{-0.3cm}
\caption{Comparison with the classic and state-of-the-art unsupervised methods on FLICKR25K in terms of mAP (\%).
}
\label{table:sota_flickr25k}
\end{table}

\begin{figure}[t]
\centering
\begin{subfigure}[]{0.47\columnwidth}{
\includegraphics[scale=0.44]{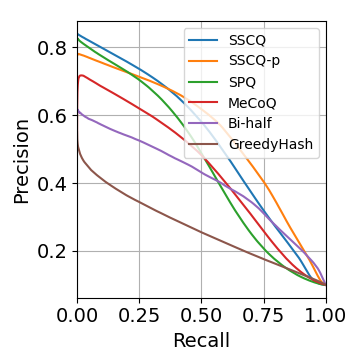}
\caption{PR curve on CIFAR10.}
\label{fig:PR_cifar}
}
\end{subfigure}
\begin{subfigure}[]{0.47\columnwidth}{
\includegraphics[scale=0.44]{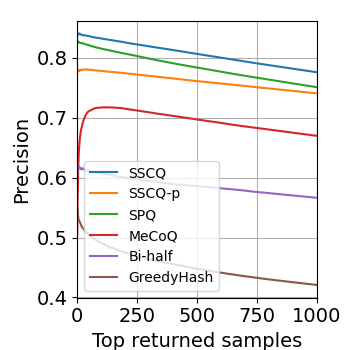}
\caption{P@1000 curve on CIFAR10.}
\label{fig:P1000_cifar}
}
\end{subfigure}
\begin{subfigure}[]{0.47\columnwidth}{
\includegraphics[scale=0.44]{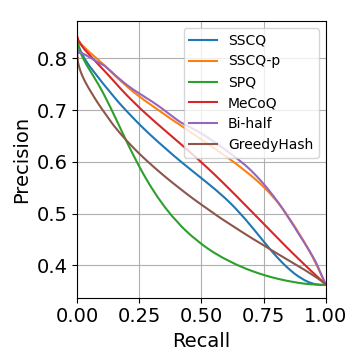}
\caption{PR curve on NUSWIDE.}
\label{fig:PR_nuswide}
}
\end{subfigure}
\begin{subfigure}[]{0.47\columnwidth}{
\includegraphics[scale=0.44]{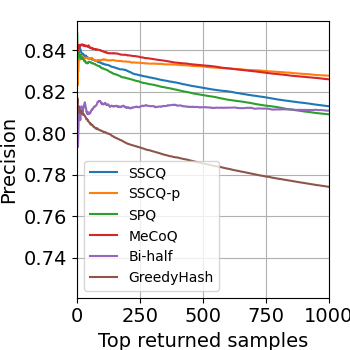}
\caption{P@1000 curve on NUSWIDE.}
\label{fig:P1000_nuswide}
}
\end{subfigure}
\begin{subfigure}[]{0.47\columnwidth}{
\includegraphics[scale=0.44]{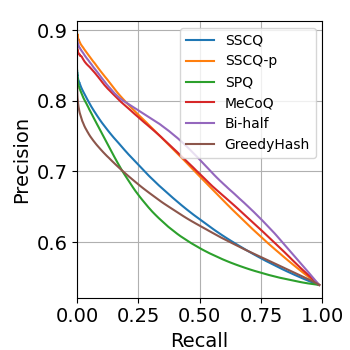}
\caption{PR curve on FLICKR.}
\label{fig:PR_flickr25k}
}
\end{subfigure}
\begin{subfigure}[]{0.47\columnwidth}{
\includegraphics[scale=0.44]{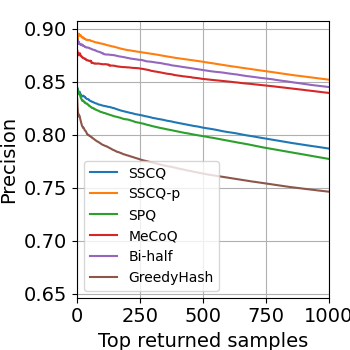}
\caption{P@1000 curve on FLICKR.}
\label{fig:P1000_flickr25k}
}
\end{subfigure}
\vspace{-0.2cm}
\caption{PR curves and P@1000 curves on CIFAR-10, NUS-WIDE and FLICKR25K (32 bits).}
\label{fig:PR_P_curve}
\end{figure}

\subsection{Comparison with the State-of-the-Arts}
We compared our SSCQ and SSCQ-p with three types of classic and state-of-the-art unsupervised image retrieval methods, including \textbf{(1)} shallow methods with input features extracted from an ImageNet pre-trained VGG16 model~\cite{simonyan2014very}, such as SpectralH~\cite{weiss2008spectral} and ITQ~\cite{gong2012iterative};
\textbf{(2)} deep pre-trained unsupervised methods which use an ImageNet pre-trained VGG16 model~\cite{simonyan2014very} as the backbone and optimize certain layers to generate codes in an unsupervised learning manner, such as Bi-half~\cite{li2021deep} and MeCoQ~\cite{wang2022contrastive};
\textbf{(3)} deep fully unsupervised methods which train a model from scratch and jointly optimize visual features and codes in an unsupervised learning manner, such as SPQ~\cite{jang2021self}.

On CIFAR-10, as shown in Table~\ref{table:sota_cifar}, our SSCQ approach achieves the best performance on all bits.
SSCQ improves the second-best SPQ~\cite{jang2021self} by 1.5\%, 2.0\% and 1.7\% on 16 bits, 32 bits and 64 bits, respectively.
Besides, with a pre-trained backbone model, our SSCQ-p approach also outperforms the state-of-the-art deep pre-trained unsupervised methods, such as Bi-half~\cite{li2021deep} and MeCoQ~\cite{wang2022contrastive}, by a distinct margin.
On the large-scale multi-label NUS-WIDE dataset, as shown in Table~\ref{table:sota_nus}, our SSCQ outperforms the state-of-the-art deep fully unsupervised SPQ by 3.0\%, 0.5\% and 0.6\% on 16 bits, 32 bits and 64 bits respectively.
And our SSCQ-p achieves 80.3\%, 81.9\% and 82.6\% on 16 bits, 32 bits and 64 bits respectively, which are competitive against the state-of-the-art deep pre-trained unsupervised MeCoQ.
On the relatively small-scale FLICKR25K dataset, as shown in Table~\ref{table:sota_flickr25k}, our SSCQ improves the state-of-the-art deep fully unsupervised SPQ approximately 2\% on all bits, while our SSCQ-p achieves 81.9\%, 82.6\% and 82.8\% on 16 bits, 32 bits and 64 bits respectively, which are still competitive against the state-of-the-art deep pre-trained unsupervised MeCoQ and Bi-half.
In addition, since NUS-WIDE and FLICKR25K are composed of natural images which are close to those from ImageNet, we can observe that SSCQ-p performs better than SSCQ, but SSCQ still yields competitive performance even though no pre-trained model is used.
This shows that our approach can accommodate to different scenarios and achieve compelling results.
Overall, the superior performance of our SSCQ and SSCQ-p can be attributed to exploring richer self-supervision to facilitate the joint optimization of visual features and codes.

Besides, in Fig.~\ref{fig:PR_P_curve}, we report PR curves and P@1000 curves.
It can be observed that our SSCQ (blue curves) consistently outperforms SPQ (green curves) under the fully unsupervised setting, while our SSCQ-p (orange curves) performs competitively against the state-of-the-art pre-trained unsupervised methods.
This further shows that our approach is capable of learning effective representations and codes for image retrieval at different required recall rates and numbers of top returned samples.

\begin{table}[t]
   \begin{center}
   \resizebox{0.99\columnwidth}{!}{
      \begin{tabular}{c|c|c|c|c|ccc}
         \hline
         \multicolumn{5}{c|}{\multirow{1}{*}{\tabincell{c}{Component}}}&\multicolumn{3}{c}{mAP (\%)}\\
         \hline
         $\mathcal{L}_{icz}$ &$\mathcal{L}_{pn}$ &$\mathcal{L}_{cd}$&$\mathcal{L}_{icf}$ &$\mathcal{L}_{cc}$  &16 bits &32 bits &64 bits \\
         \hline
         \hline
         $\checkmark$ &- &- &- &- &74.2 &77.6 &78.5 \\
         $\checkmark$ &$\checkmark$ &- &- &-  &77.3 &79.2 &80.8 \\
         $\checkmark$ &$\checkmark$ &$\checkmark$ &- &- &77.9 &80.6 & 81.9 \\
         $\checkmark$ &- &- &$\checkmark$ &- &76.5 &80.0 &80.8 \\
         $\checkmark$ &- &- &$\checkmark$ &$\checkmark$ &76.8 &80.2 &81.4 \\
         $\checkmark$ &$\checkmark$ &$\checkmark$ &$\checkmark$ &$\checkmark$ &\textbf{78.3} &\textbf{81.3} &\textbf{82.9} \\
         \hline
      \end{tabular}
   }
   \end{center}
\vspace{-0.3cm}
\caption{Component effectiveness evaluation on CIFAR-10.
}
\label{table:component}
\end{table}

\subsection{Ablation Study}
In Table~\ref{table:component}, we present component effectiveness evaluation of the proposed SSCQ on CIFAR-10.
We can observe that:
\textbf{(1)} While the baseline model ($\mathcal{L}_{icz}$) achieves 74.2\% on 16 bits, 77.6\% on 32 bits and 78.5\% on 64 bits, the part consistent quantization ($\mathcal{L}_{icz}$+$\mathcal{L}_{pn}$+$\mathcal{L}_{cd}$) improves the baseline by 77.9\% on 16 bits, 80.6\% on 32 bits and 81.9\% on 64 bits, respectively.
This shows that discovering part neighbor semantic structure information as self-supervision can improve contrastive quantization.
\textbf{(2)} The global consistent quantization ($\mathcal{L}_{icz}$+$\mathcal{L}_{icf}$+$\mathcal{L}_{cc}$) also significantly improves the baseline, achieving 76.8\% on 16 bits, 80.2\% on 32 bits and 81.4\% on 64 bits, respectively.
This shows the effectiveness of learning the affinity between instances as richer self-supervision for contrastive quantization.
\textbf{(3)} SSCQ with the unified learning objective ($\mathcal{L}_{icz}$+$\mathcal{L}_{icf}$+$\mathcal{L}_{cc}$+$\mathcal{L}_{pn}$+$\mathcal{L}_{cd}$) of global and part consistent quantization yields the best performance, namely 78.3\% on 16 bits, 81.3\% on 32 bits and 82.9\% on 64 bits, which improves the baseline by 4.1\% on 16 bits, 3.7\% on 32 bits, 4.4\% on 64 bits, respectively.
These results validate that exploring richer part and global self-supervision in contrastive quantization can facilitate model optimization to obtain better performance.

\begin{figure}[t]
\centering
\begin{subfigure}[]{0.49\linewidth}{
\includegraphics[scale=0.42]{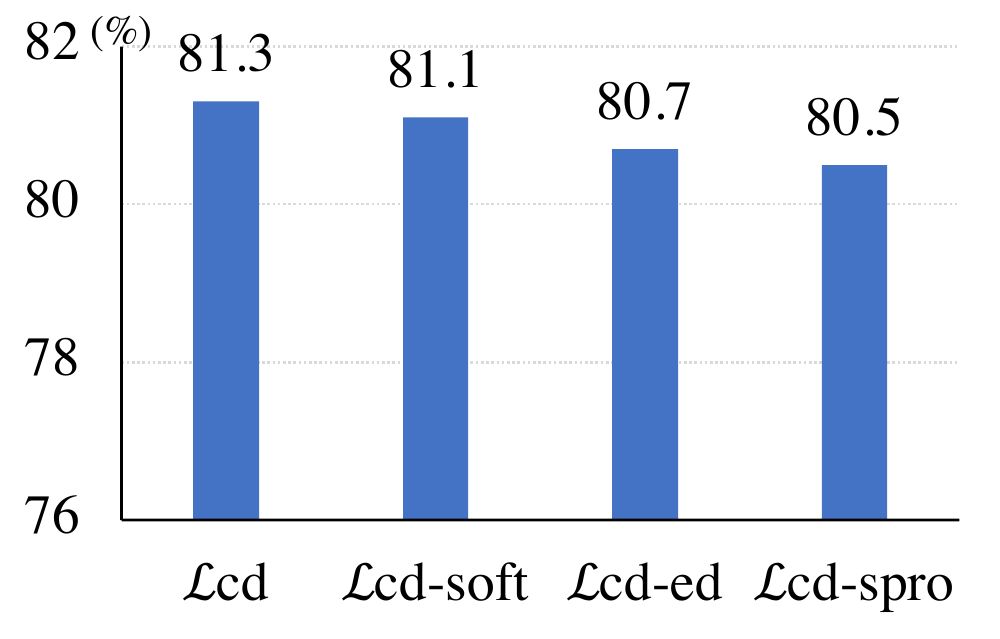}
\caption{$\mathcal{L}_{cd}$ variants.}
\label{fig:diversity}
}
\end{subfigure}
\begin{subfigure}[]{0.49\linewidth}{
\includegraphics[scale=0.42]{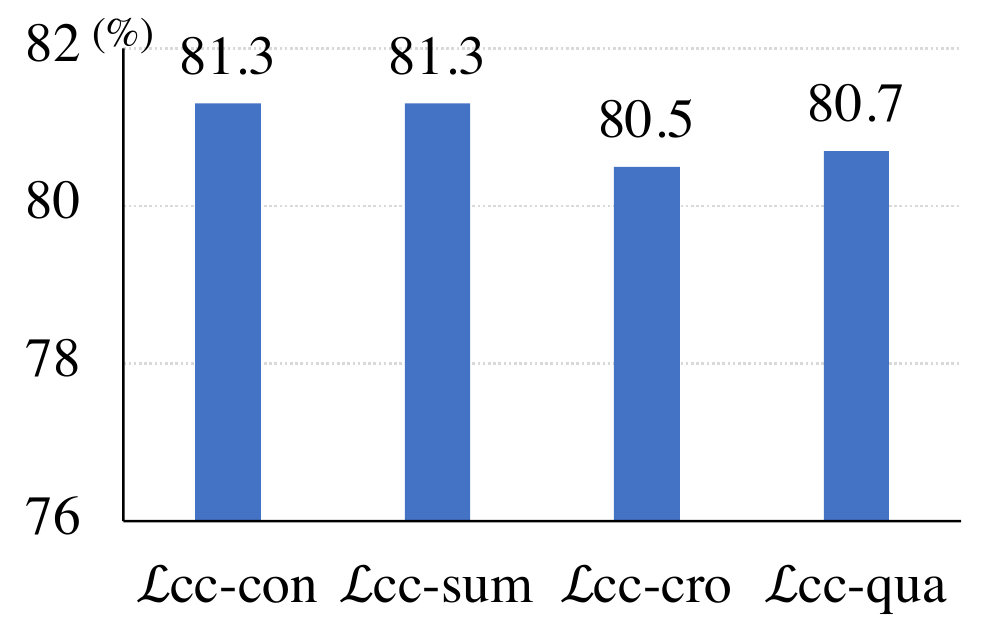}
\caption{Fusion variants.}
\label{fig:fusion}
}
\end{subfigure}
\vspace{-0.2cm}
\caption{Evaluating (a) codeword diversity regularization variants and (b) representation fusion variants on CIFAR-10 (32 bits).}
\end{figure}

\subsection{Further Analysis and Discussion}
\noindent\textbf{Codeword Diversity Regularization Variants.}
As shown in Fig.~\ref{fig:diversity}, we test SSCQ with different codeword distribution regularization strategies, where $\mathcal{L}_{cd-soft}$ and $\mathcal{L}_{cd-ed}$ denote soft quantization based or squared Euclidean distance based probability to compute the mean output probability in Eq.\eqref{eq:codeword_diversity}, and $\mathcal{L}_{cd-spro}$ denotes using squared probability as \cite{klein2019end} in Eq.\eqref{eq:codeword_diversity}.
Overall, SSCQ with entropy maximization based codeword diversity regularization achieves encouraging result.

\vspace{0.2cm}
\noindent\textbf{Representation Fusion Variants.}
In Fig.~\ref{fig:fusion}, we report the performance of SSCQ with different embedding and quantized representations fusion strategies, including concatenation fusion, sum fusion, cross consistent contrastive regularization, as well as directly computing $\mathcal{L}_{cc}$ with quantized representations.
We can observe that SSCQ with $\mathcal{L}_{cc-con}$ and $\mathcal{L}_{cc-sum}$ yield compelling results, while SSCQ with $\mathcal{L}_{cc-cro}$ and $\mathcal{L}_{cc-qua}$ as \cite{wei2021co2} achieve slightly worse results.

\begin{figure}[t]
\centering
   \includegraphics[width=0.8\linewidth]{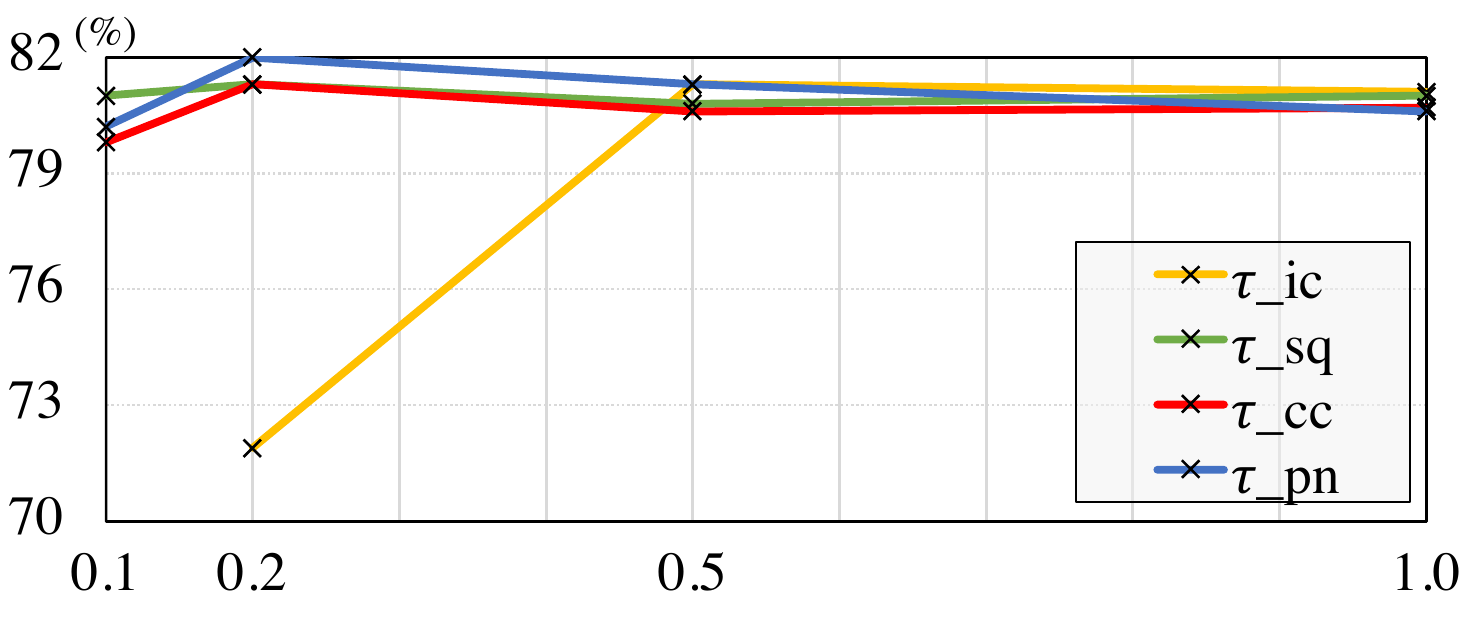}
\vspace{-0.2cm}
\caption{Evaluating temperature parameter sensitivity on CIFAR-10 (32 bits).}
\label{fig:temperature}
\end{figure}

\vspace{0.2cm}
\noindent\textbf{Temperature Parameter Sensitivity.}
In Fig.~\ref{fig:temperature}, we report the performance of our SSCQ with the temperature parameters ($\tau_{ic}$, $\tau_{sq}$, $\tau_{cc}$ and $\tau_{pn}$) ranging from different values.
It can be seen that with different values of $\tau_{cc}$, $\tau_{pn}$ and $\tau_{sq}$, SSCQ still performs competitively and the reasonable range is at around $0.2$ for $\tau_{cc}$, $[0.2, 0.5]$ for $\tau_{pn}$ and $\tau_{sq}$.
As for $\tau_{ic}$, it is a basic component of contrastive quantization, so it is more sensitive and yields competitive result at around $[0.5, 1.0]$.

\begin{figure}[t]
\centering
   \includegraphics[width=0.99\linewidth]{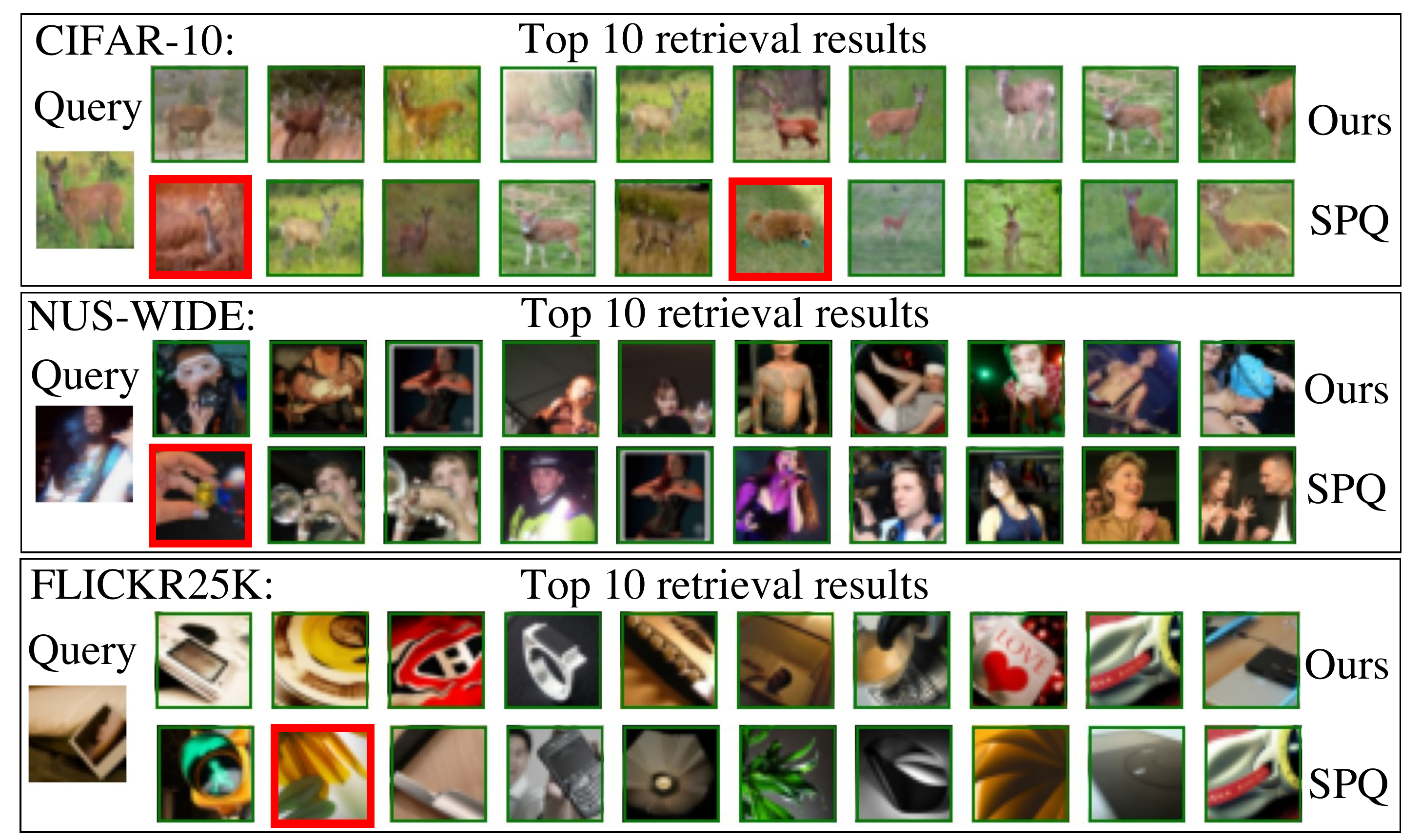}
\vspace{-0.2cm}
\caption{Retrieval results of our approach and SPQ on CIFAR-10, NUS-WIDE and FLICKR25K (32 bits). False retrieval results are in red bounding boxes.}
\label{fig:vis_qualitative}
\end{figure}

\vspace{0.2cm}
\noindent\textbf{Qualitative Retrieval Results.}
In Fig.~\ref{fig:vis_qualitative}, we visualize some retrieval results returned by our SSCQ and SPQ~\cite{jang2021self} respectively.
We can see that both SSCQ and SPQ can retrieve visually similar images from the database, but SSCQ tends to explore more discriminative information resulting in more relevant retrieval results.
This mainly attributed to discovering richer self-supervision for model learning in SSCQ.

\section{Conclusion}
In this work, we propose a novel Self-Supervised Consistent Quantization (SSCQ) approach to deep fully unsupervised image retrieval.
To discover underlying semantic structure information as self-supervision, we devise \emph{part consistent quantization} using part neighbor semantic consistency learning with codeword regularization and \emph{global consistent quantization} using instance contrastive learning with consistent contrastive regularization.
We formulate the part and global consistent quantization into a unified learning objective to facilitate model learning.
Extensive experiments on three benchmark datasets show the superior performance of our approach over the state-of-the-art methods.
In future work, we aim to explore richer self-supervision information to facilitate the joint learning of visual features and codes for more efficient image retrieval.

{\small
\bibliographystyle{ieee_fullname}
\bibliography{reference}
}

\end{document}